%% file: main.tex
\documentclass{article} 
\usepackage{_ICLR_package/iclr2027_conference,times} 

\input{_ICLR_package/math_commands}

\usepackage{hyperref}
\usepackage{url}
\usepackage{xspace}
\usepackage{wrapfig}
\usepackage{tabularx}
\usepackage{array}
\usepackage{algpseudocode}
\usepackage{algorithm}

\usepackage{booktabs}
\usepackage{multirow}
\usepackage{amsfonts}
\usepackage{graphicx}  

\usepackage{amssymb}
\usepackage{colortbl}
\usepackage{pifont}
\usepackage{makecell}
\usepackage{enumitem} 
\usepackage{adjustbox}
\usepackage{pifont}     
\usepackage{xcolor}     
\usepackage{tikz}
\usetikzlibrary{backgrounds}

\title{\proposed: A Unified Interface for Agentic Web Access}

\author{Sunghwan Kim, Deogyong Kim, Sangam Lee, Donguk Kwon, Dongha Lee\footnotemark[2] \\
Yonsei University\\
\texttt{\{happysnail06, legenduck, salee, donguk.kwon, donalee\}@yonsei.ac.kr} \\
}

\iclrfinalcopy 

\begin{document}
\input{dfn}

\maketitle
\begin{abstract}
\input{MainText/000_abstract}
\end{abstract}


\section{Introduction}
\label{sec:introduction}
\input{MainText/010_introduction}

\section{Related Work}
\label{sec:relatework}
\input{MainText/020_relatedwork}

\section{Preliminary Analysis: Can agents effectively access the web through existing access models?}
\label{sec:preliminary_analysis}
\input{MainText/030_preliminary_analysis}

\section{\proposed}
\label{sec:method}
\input{MainText/040_method}

\section{Experiments}
\label{sec:experiment}
\input{MainText/050_experiments}

\section{Analysis}
\label{sec:analysis}
\input{MainText/060_analysis}

\section{Conclusion}
\label{sec:conclusion}
\input{MainText/070_conclusion}

\input{MainText/090_statements}
\bibliography{bibliography}
\bibliographystyle{_ICLR_package/iclr2027_conference}


\appendix
\input{MainText/080_appendix}

\end{document}

%% file: _ICLR_package/math_commands.tex
\usepackage{amsmath,amsfonts,bm}

\def\eqref#1{equation~\ref{#1}}

\def\1{\bm{1}}

\DeclareMathAlphabet{\mathsfit}{\encodingdefault}{\sfdefault}{m}{sl}
\SetMathAlphabet{\mathsfit}{bold}{\encodingdefault}{\sfdefault}{bx}{n}



%% file: dfn.tex
\newcommand{\proposed}{\textsc{Web\$hell}\xspace}
\newcommand{\webcontract}{\textsc{WebContract}\xspace}
\newcommand{\webcompiler}{\textsc{WebCompiler}\xspace}
\newcommand{\webcache}{\textsc{WebCache}\xspace}
\newcommand{\hybridsb}{Hybrid\textsubscript{SB}\xspace}
\newcommand{\hybridsbf}{Hybrid\textsubscript{SBF}\xspace}

%% file: MainText/000_abstract.tex
Large language model (LLM) agents are becoming direct users of the open web. As they take on increasingly complex and information-intensive tasks, effective web access is becoming a core capability for agentic problem solving.
Existing approaches have advanced agentic web access through different \textit{access models}, including search-augmented agents, browser-based agents, and, more recently, filesystem-based agents.
Nevertheless, as users expect agents to complete long-horizon, end-to-end tasks, they increasingly need to conduct broad information gathering, perform browser actions, and revisit acquired evidence within a single trajectory.
Yet our preliminary analyses reveal that existing access models struggle to meet these demands. They primarily support either information gathering or browser interaction, while combining their capabilities can involve substantial costs or performance trade-offs.
To this end, we present \proposed, an interface that unifies web access under a single access model. Using a common schema defined by \webcontract, \webcompiler compiles live webpages into addressable information and action \textit{objects}.
These objects are maintained in \webcache and handled through a shared set of operations, providing agents with consistent, operable, and persistent web access.
Through extensive experiments across web search, web navigation, and deep research benchmarks, we show that \proposed achieves strong task performance at lower cost.
Our results further indicate the importance of interface design for effective web access, with \proposed opening new possibilities for more capable and efficient LLM agents on the web.

%% file: MainText/010_introduction.tex
Large language model agents are becoming direct users of the open web.
As they take on increasingly information-intensive and open-ended tasks, effective web access is becoming a core capability for agentic problem solving~\citep{deng2023mind2web, kim2026agenticshop}.
Yet making effective use of the web is far from trivial, requiring substantial reasoning to locate information and interact with heterogeneous webpages.
This raises a fundamental question of how agents should access the web.

Existing research has approached this problem through different \textit{access models}.
Search-augmented agents~\citep{li2025searcho1, jin2025searchr, zheng-etal-2025-deepresearcher} operate through search APIs and cleaned text, while browser-based agents~\citep{Zhou2024webarena, yang2025agentoccam,chae2025web} operate through page observations and actions.
More recently, filesystem-based methods~\citep{liu2026fetch} have enabled agents to store web documents and selectively inspect their contents through shell operations (e.g., \texttt{grep}, \texttt{read}), reducing the need to read the entire contents.
For the most part, these lines of work have made substantial progress within their respective access models.
However, as users expect agents to complete long-horizon, end-to-end tasks, agents are increasingly required to perform multiple forms of web access within a single trajectory.
For example, a shopping query may require an agent to discover relevant products, navigate pagination to find reviews, and retain useful information for later comparison.
Yet these capabilities have largely been developed through separate access models, motivating a rethinking of how they can be integrated for agentic web access.

Accordingly, this paper begins by investigating how well existing access models generalize across web search and web navigation, two representative types of web tasks.
The key bottleneck is that each access model supports certain capabilities at the expense of others.
We find that (i) search-augmented and filesystem-based agents struggle with web navigation since clean-text representations often omit page controls, while browser-based agents are less effective at broad information gathering due to their lengthy, noisy observations; (ii) combining search and browser-based access broadens task coverage but incurs substantial costs from repeated webpage interpretation; and (iii) adding filesystem-based capabilities such as \texttt{grep} reduces these costs.
However, selective inspection can miss the context needed to use page controls, degrading web navigation performance.

Together, these findings highlight three requirements for integrating the capabilities of existing access models.
First, \textit{\textbf{operability}} requires supporting both efficient information gathering and effective webpage interaction.
Second, \textit{\textbf{consistency}} calls for unifying webpage representations under common schemas, reducing the need to reinterpret the same page.
Finally, \textit{\textbf{persistence}} requires retaining retrieved web documents for selective access, lowering cost while preserving operability.

To this end, we present \proposed, an interface that unifies web access under a single access model.
Our key idea is to provide agents with an intermediary that standardizes webpage representations and browser actions.
Instead of coordinating separate access models, agents interact with \proposed through a request-response loop.
Specifically, \webcontract defines how information and available actions are represented as objects, organized into pages, and accessed through shared operations.
Each object combines a \textit{value}, an \textit{address}, and \textit{associations} that preserve its context and references to related objects.
Following this contract, \webcompiler transforms live webpages into structured representations using deterministic rules derived from HTML semantics.
\webcache then retains the compiled pages for selective access and reuse throughout a task.
Importantly, \proposed allows agents to use object addresses to specify which information to inspect or which action to execute in subsequent requests.
The interface then carries out the requested operation on the agent's behalf and returns the outcome, thereby enabling unified agentic web access.

We conduct extensive experiments on six benchmarks spanning web search, web navigation, and deep research.
Notably, \proposed achieves the highest task performance on all six benchmarks at comparable or lower cost.
Our analyses suggest that an effective web interface should make clear how webpage elements relate to one another.
These connections help agents gather more evidence, make browser actions more intuitive, and potentially guide exploration beyond the content already inspected.
We further show that \proposed remains robust to irrelevant content and maintains its efficiency as retrieved documents accumulate.
Together, these findings demonstrate \proposed's effectiveness across complex, information-intensive web tasks.\
Our contributions are as follows:
\begin{itemize}[leftmargin=*,topsep=2pt,itemsep=2pt,parsep=0pt]
  \item We identify the need for unified agentic web access, investigating the limitations of existing access models and the challenges of integrating their capabilities.
  
  \item We propose \proposed, an interface that provides consistent, operable, and persistent web access, enabling agents to handle diverse web tasks under a single access model.
  
  \item We demonstrate that \proposed achieves strong task performance and high efficiency across diverse web tasks, providing insights to guide future research on agentic web access.
\end{itemize}

\vspace{-0.2cm}

%% file: MainText/020_relatedwork.tex
\paragraph{Agents for Web.}
Research on web-capable agents has advanced both information seeking and web navigation.
Search-augmented agents interleave web search with reasoning~\citep{li2025searcho1,jin2025searchr}, while browser-based agents use page observations to guide their actions~\citep{he2024webvoyager,Zheng2024seeact}.
More recently, inspired by direct corpus interaction~\citep{li2026beyond,zhuang2026rise,lu2026drdci}, a complementary line of research has explored filesystem-based web access, where webpages are fetched into persistent workspaces and selectively inspected with shell operations~\citep{liu2026fetch}.
Open-source software such as CLI-Anything Browser~\citep{yang2026clianything} has also adopted a similar approach, supporting web navigation by applying shell operations to the accessibility tree of a webpage.
Nevertheless, these lines of work have largely developed web search and web navigation within separate access models.
\proposed unifies their capabilities at the interface level, standardizing web access through common representations and operations.

\vspace{-0.1cm}
\paragraph{Interface for Agents.}
Recent work on agent-computer interfaces~\citep{yang2024sweagent} and model harnesses ~\citep{lee2026metaharness, wang2026harnessbridge, jiang2026harness1} has established interface design as an important research problem for improving agent performance.
On the web, this perspective initially took the form of simplifying webpage representations, including element filtering and DOM pruning~\citep{yang2025agentoccam,Zhang2026prune4web}.
Subsequent work~\citep{lee2025learning, kwon2026region4web,hwang2026webchallenger} has moved toward enhancing page representations, organizing webpage elements into regions or summaries that support targeted interaction.
Yet these approaches require additional model inference or extensive offline traversal, limiting their practicality for open-web access.
Alternatively, WebMCP and NLWeb~\citep{steiner2026mcp}  place the responsibility for interface design on website providers, who encapsulate site-specific functions and content for agents.
These works share the premise that interfaces should help agents access and use the web~\citep{zhong2026agentwebbench}, but each applies this premise within a limited scope.
\proposed extends it toward general web access, providing agents with an operable, consistent, and persistent interface.

\vspace{-0.2cm}

%% file: MainText/030_preliminary_analysis.tex
\input{Tables/030_access_models_table}
To understand how web access should be designed for LLM agents, we first analyze the capabilities and limitations of existing access models.
Specifically, our analysis addresses two questions:

\textbf{Question I:} How well do existing access models generalize across web tasks?

\textbf{Question II:} What limitations remain when access models are combined?

\textbf{Benchmarks.}
We consider two types of web tasks: web search and web navigation.
We use WideSearch~\citep{wong2026widesearch} for web search, which requires broad information gathering to populate tables, and WebArena-Verified~\citep{webarena-verified} for web navigation, which requires deeper, multi-step website interaction (e.g., reordering products).
We randomly sample 100 queries from each benchmark and report item-level F1 for WideSearch and success rate for WebArena-Verified.

\paragraph{Access Models.}
We evaluate three families of agents with different web access models (Table~\ref{tab:access_model_comparison}).
\begin{itemize}[leftmargin=1.5em,topsep=0pt,partopsep=0pt,itemsep=3pt,parsep=0pt]
\par\vspace{-3pt}
\item \textbf{\textit{Search-augmented Agents}}.
These agents primarily address web search tasks using search APIs and clean-text representations of webpages.
Fetch-and-Read directly reads webpages returned by a web search.
Search-o1~\citep{li2025searcho1} uses its Reason-in-Documents module to summarize relevant information from these pages, while WebThinker~\citep{li2025webthinker} uses an explorer agent to browse deeper into websites for additional information.
\item \textbf{\textit{Filesystem-based Agents}}.
Fetch-then-Explore (FtE)~\citep{liu2026fetch} also targets web search tasks, but stores cleaned webpage text in a filesystem workspace.
Agents can then selectively locate and read relevant information using shell operations such as \texttt{grep} and \texttt{read}.
\item \textbf{\textit{Browser-based Agents}}.
These agents are primarily designed for web navigation tasks, using structured page observations such as accessibility trees (AXTrees) to guide browser interactions.
We include AgentOccam~\citep{yang2025agentoccam}, Agent-E~\citep{agente}, and BrowserUse~\citep{browser_use2024} as representative browser-based agents.
\end{itemize}
\par\vspace{-3pt}
For our analysis,  we refer to search-augmented agents and Fetch-then-Explore as \textit{search-oriented}, and web navigation agents as \textit{navigation-oriented}.
For evaluation on WebArena-Verified, we augment the search-oriented agents with the browser actions introduced in WebArena~\citep{Zhou2024webarena}, while retaining their clean-text observations.
We use GPT-5.6-Luna~\citep{openai2026gpt56} as the backbone for all agents (reasoning: \textit{high}).
Further details are provided in Appendix~\ref{appendix_preliminary_analysis_details}.

%
\subsection{Analysis I: Can Existing Access Models Generalize Across Web Tasks?}
\label{sec:preliminary_analysis_1}
\paragraph{Existing access models struggle beyond their primary tasks.}
We report cross-task performance of navigation-oriented agents on WideSearch and search-oriented agents on WebArena-Verified, using the other group's average performance on each benchmark as a reference (Figure~\ref{fig:preliminary_cross_task}).
\input{Figures/preliminary_analysis/1_preliminary_cross_task_figure}
We find that agents consistently fall below the corresponding reference average outside their primary task setting.
These cross-task performance gaps can be partly explained by how each access model presents webpages to the agent.
Clean-text observations are well suited for gathering information across many pages.
However, text extraction can discard the structure and attributes needed to use controls such as buttons and input fields, limiting support for browser interaction.
Although URL-based fetching supports limited navigation by allowing agents to follow links, it does not enable them to operate these controls.
Browser observations preserve interactive elements, but their length and noise can consume substantial context when gathering information across many sources.

\subsection{Analysis II: What limitations remain when access models are combined?}
\label{sec:preliminary_analysis_2}
\paragraph{Combining search and browser-based access supports both web tasks, but incurs substantial cost.}
\input{Tables/030_preliminary_hybrid_cost_table}
Motivated by the findings in Analysis I, we first examine whether a hybrid access model can support both web search and web navigation tasks.
We introduce \hybridsb (\textit{Search + Browser}), which provides clean text for information gathering and AXTrees with browsing tools for interaction.
\hybridsb performs comparably to the search-oriented average on WideSearch (66.0\% vs.\ 65.8\%) and exceeds the navigation-oriented average on WebArena-Verified (74.0\% vs.\ 54.7\%).
However, it incurs substantially higher input token usage and LLM cost (Table~\ref{tab:preliminary_call_cost}).
We observe that \hybridsb repeatedly accesses the same source through both clean text and AXTrees for two purposes:
(1) when the desired information is missing or unclear in one representation, agents use the other to find it.
(2) agents cross-check page representations to verify their completeness.
These repeated checks significantly increase the cost, indicating the need for a unified access model that reduces reliance on separate representations.

\paragraph{Adding filesystem access with shell operations creates a tradeoff between web search and web navigation.}
\input{Figures/preliminary_analysis/2_preliminary_hybrid_figure}
We next examine \hybridsbf, which extends \hybridsb with a filesystem workspace and shell operations.
We observe that \hybridsbf performs better at lower cost on WideSearch, where tasks require broad source coverage but only specific evidence from each source (Figure~\ref{fig:preliminary_hybrid}, left).
However, performance declines on WebArena-Verified (Figure~\ref{fig:preliminary_hybrid}, right), where completing tasks requires identifying and using page controls.
Selective inspection may reveal relevant text without showing how to interact with the page, causing agents to exhaust their turn budget searching for the required controls.

\subsection{Insights from Preliminary Analysis}
Our analyses identify three requirements for agentic web access.
\textbf{Operability} preserves the context needed for browser interaction while enabling efficient information gathering.
\textbf{Consistency} calls for a common representation of information and browser actions, reducing repeated webpage interpretation.
\textbf{Persistence} retains retrieved web documents for selective access, reducing the need to read pages in full.
Together, these requirements guide the design of a unified web access model.

%% file: Tables/030_access_models_table.tex

\definecolor{darkgreen}{HTML}{006400}
\newcommand{\cmark}{\textcolor{darkgreen}{\ding{51}}}
\newcommand{\xmark}{\textcolor{red}{\ding{55}}}

\begin{wraptable}{r}{0.55\textwidth}
\centering
\vspace{-2.2em}
\caption{Overview of different web access models.}
\vspace{0.2cm}
\label{tab:access_model_comparison}
\setlength{\tabcolsep}{5pt}
\renewcommand{\arraystretch}{1.1}

\resizebox{\linewidth}{!}{%
\begin{tabular}{lccccc}
\toprule
\multirow{2}{*}{\textbf{Method}}
& \multicolumn{2}{c}{\textbf{Representation}}
& \multicolumn{3}{c}{\textbf{Interface}} \\
\cmidrule(lr){2-3} \cmidrule(lr){4-6}
& \textbf{Clean Text}
& \textbf{AXTree}
& \textbf{Search API}
& \textbf{Browser}
& \textbf{Filesystem} \\
\midrule

\multicolumn{6}{l}{\textbf{\textit{Search-augmented}}} \\
Fetch-and-Read
& \cmark & \xmark & \cmark & \xmark & \xmark \\
Search-o1
& \cmark & \xmark & \cmark & \xmark & \xmark \\
WebThinker
& \cmark & \xmark & \cmark & \xmark & \xmark \\

\midrule
\multicolumn{6}{l}{\textbf{\textit{Browser-based}}} \\
Agent-E
& \xmark & \cmark & \xmark & \cmark & \xmark \\
AgentOccam
& \xmark & \cmark & \xmark & \cmark & \xmark \\
BrowserUse
& \xmark & \cmark & \xmark & \cmark & \xmark \\

\midrule
\multicolumn{6}{l}{\textbf{\textit{Filesystem-based}}} \\
Fetch-then-Explore
& \cmark & \xmark & \cmark & \xmark & \cmark \\

\midrule
\multicolumn{6}{l}{\textbf{\textit{Hybrid Config.}}} \\
Hybrid\textsubscript{SB}
& \cmark & \cmark & \cmark & \cmark & \xmark \\
Hybrid\textsubscript{SBF}
& \cmark & \cmark & \cmark & \cmark & \cmark \\

\midrule
\textbf{\proposed (ours)}
& \multicolumn{2}{c}{\textbf{Addressable Objects}}
& \multicolumn{3}{c}{\textbf{Unified Interface}} \\

\bottomrule
\end{tabular}%
}
\vspace{-1.8em}
\end{wraptable}

%% file: Figures/preliminary_analysis/1_preliminary_cross_task_figure.tex
\begin{wrapfigure}{r}{0.51\textwidth}
    \centering
    \includegraphics[width=\linewidth]{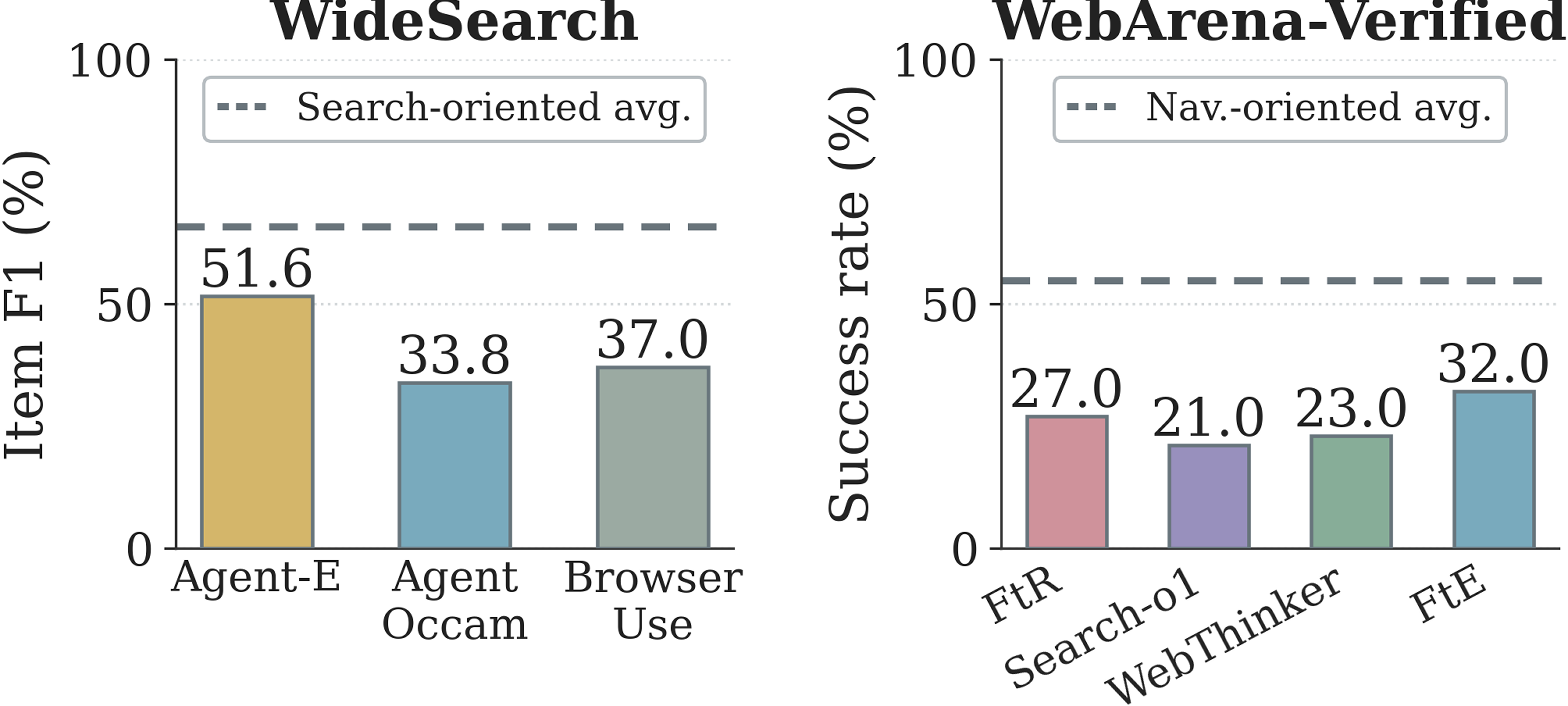}
    \vspace{-1.6em}
    \caption{
        Cross-task performance of navigation-oriented agents on WideSearch and search-oriented agents on WebArena-Verified. Dashed lines indicate the average performance of search-oriented and navigation-oriented agents, respectively.
    }
    \vspace{-0.75em}
    \label{fig:preliminary_cross_task}
\end{wrapfigure}

%% file: Tables/030_preliminary_hybrid_cost_table.tex
\definecolor{pct-increase}{RGB}{224,79,80}
\definecolor{pct-baseline}{RGB}{150,150,150}

\begin{wraptable}{r}{0.4\textwidth}
\vspace{-2.2em}
\centering
\small
\setlength{\tabcolsep}{2.5pt}
\caption{Mean input tokens (k/call) and LLM cost (USD/100 calls). $\Delta$: relative increase over reference averages.}
\vspace{3pt}
\label{tab:preliminary_call_cost}
\resizebox{\linewidth}{!}{%
\begin{tabular}{@{}l*{2}{rr}}
\toprule
\multirow{2}{*}{\textbf{Agent}}
& \multicolumn{2}{c}{\textbf{Input tokens}}
& \multicolumn{2}{c}{\textbf{LLM cost}} \\
\cmidrule(lr){2-3} \cmidrule(lr){4-5}
& Avg. & $\Delta$ & Avg. & $\Delta$ \\
\midrule
\multicolumn{5}{@{}l}{\textit{WideSearch}} \\
\addlinespace[2pt]
\quad Search avg.
& 46.7 & \textcolor{pct-baseline}{-{}-\%} & 0.33 & \textcolor{pct-baseline}{-{}-\%} \\
\quad \hybridsb
& 182.2 & \textcolor{pct-increase}{+290\%} & 1.21 & \textcolor{pct-increase}{+272\%} \\
\midrule
\multicolumn{5}{@{}l}{\textit{WA-verified}} \\
\addlinespace[2pt]
\quad Nav. avg.
& 10.2 & \textcolor{pct-baseline}{-{}-\%} & 0.19 & \textcolor{pct-baseline}{-{}-\%} \\
\quad \hybridsb
& 74.2 & \textcolor{pct-increase}{+626\%} & 0.39 & \textcolor{pct-increase}{+103\%} \\
\bottomrule
\end{tabular}
}
\vspace{-1.0em}
\end{wraptable}

%% file: Figures/preliminary_analysis/2_preliminary_hybrid_figure.tex
\begin{wrapfigure}{r}{0.49\textwidth}
    \centering
    \vspace{-1em}
    \includegraphics[width=\linewidth]{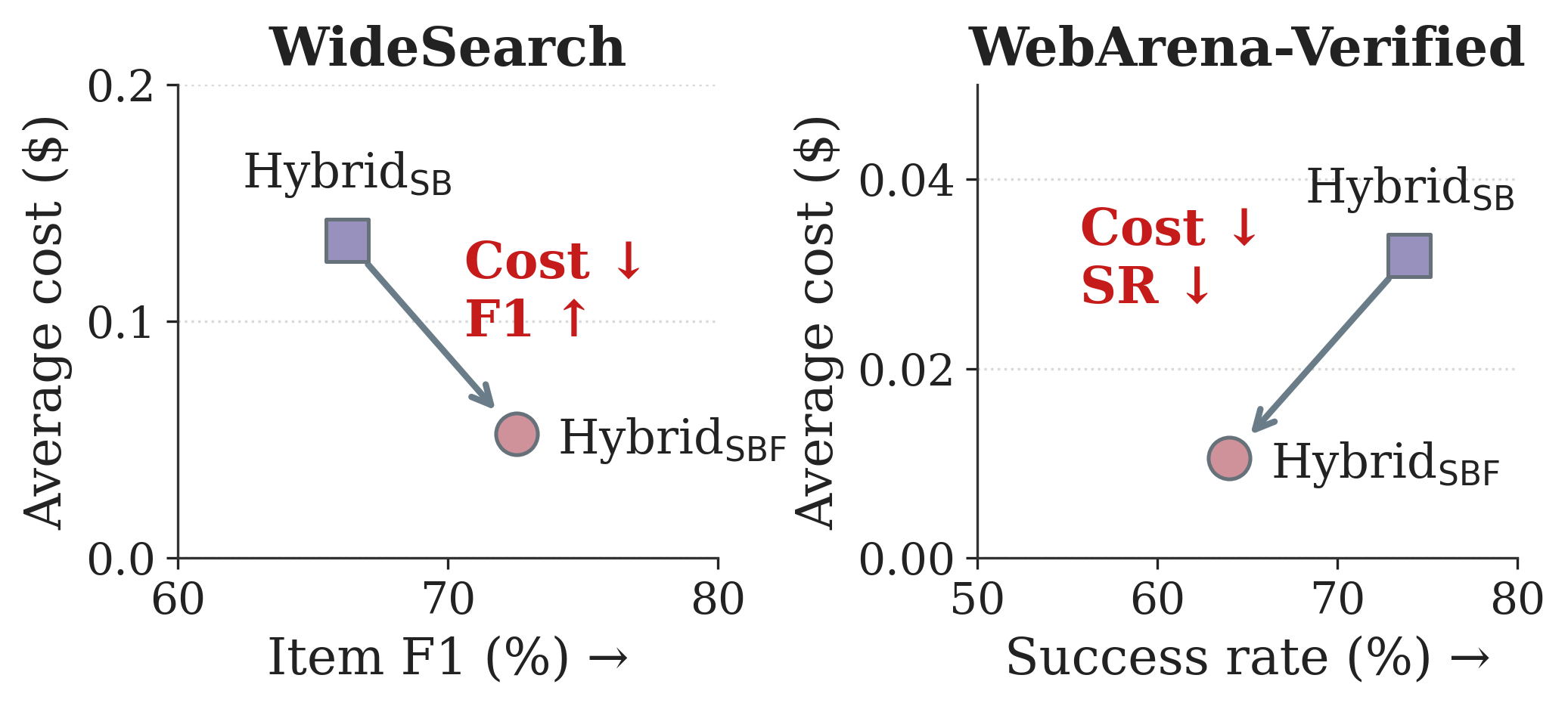}
    \vspace{-2em}
    \caption{
        Performance and cost of hybrid configurations on WideSearch and WA-Verified.
    }
    \vspace{-0.8em}
    \label{fig:preliminary_hybrid}
\end{wrapfigure}

%% file: MainText/040_method.tex
To this end, we introduce \proposed, an interface for agents that unifies web access under a single access model.
Specifically, our approach builds on three core components.
(1) \webcontract establishes common rules for representing and accessing webpages across websites.
(2) \webcompiler compiles fetched webpages into operable representations that follow these rules.
(3) \webcache preserves compiled webpages for selective and continued access throughout the task.

\subsection{Interaction Loop}
Given a user query, the agent interacts with \proposed through the \textit{Request--Response} loop illustrated in Figure~\ref{fig:main_fig}.
Requests and responses follow the common rules defined by \webcontract.
\par\vspace{-3pt}
\textbf{Agent side.}
At each step, the agent considers its current context to determine what information or interaction it needs.
It selects an operation (e.g., \texttt{retrieve}, \texttt{grep}, \texttt{execute}) and sends a request.
After receiving the response, it chooses its next request or produces a final answer.
\par\vspace{-3pt}
\textbf{Interface side.}
The interface carries out the requested operation on the agent's behalf.
For \texttt{retrieve}, it searches the web, fetches the top-ranked webpages, and captures their current states.
\webcompiler then compiles each captured webpage into \textit{information objects} that represent its content and \textit{action objects} that describe its available browser actions.
It organizes these objects into a common file structure following \webcontract, providing the agent with a consistent view across websites.
Each compiled page is assigned a page address and stored in \webcache for subsequent requests.
For \texttt{grep}, the interface locates matching content within these cached pages.
For \texttt{execute}, it invokes the specified action object to perform the corresponding interaction on the live webpage.
Updated page states are compiled and reflected in \webcache, while earlier states are kept read-only.
The interface returns the requested information or execution outcome in its response.

\input{Figures/main_figure}

\subsection{\webcontract: Standardizing Web Access}

\paragraph{Object Schema.}
The first step in designing \proposed is to determine what \textit{objects} agents should operate on.
Webpages encode content and controls in individual elements, but the context needed to interpret or use them may be encoded in separate elements.
For example, a price entry is only meaningful together with the product name and option label, which may appear in separate elements.
We therefore define an \textit{object} as a unit of information or action with the following three properties.
\begin{itemize}[leftmargin=1.5em,topsep=0pt,partopsep=0pt,itemsep=3pt,parsep=0pt]
\vspace{-4pt}
\item \textbf{Address.} A stable reference through which the agent can identify the object.
\item \textbf{Value.} The information or action represented by the object.
\item \textbf{Association.} The context needed to interpret the value, including references to other objects.
\end{itemize}
\vspace{-4pt}
This shared schema provides a consistent, addressable, and directly interpretable unit of interaction.

\vspace{-5pt}
\paragraph{Page Schema.}
The page schema defines a common file structure for representing a webpage.
A unique \textit{page address} identifies each fetched webpage as a whole.
Each object has a unique local identifier within its page.
This identifier is further combined with the page address to form its \textit{object address}.
Within each page, objects are organized into groups using a finite set of types, such as text, tables, links, and actions, grounded in HTML semantics.
Text objects are further organized into sections based on the webpage's heading structure.
A table of contents lists the available object groups and shows how to access them.
The detailed page schema is presented in Appendix~\ref{appendix_page_schema}.

\vspace{-4pt}
\paragraph{Operation Schema.}
The operation schema defines how agents communicate with the interface through \textit{requests} and \textit{responses}.
Information and action objects serve as the basic units of operation.
An agent-to-interface request specifies an operation type, its arguments, and an object address when needed.
An interface-to-agent response returns the outcome along with any resulting content and references.
These common rules apply to operations such as \texttt{retrieve}, \texttt{grep}, \texttt{read}, and \texttt{execute}.
The detailed operation schema and the list of operations are provided in Appendix~\ref{appendix_operations}.
\vspace{-4pt}

\subsection{\webcompiler: Compiling an Operable Observation Space}

\paragraph{Object Compilation.}
\webcompiler applies deterministic rules derived from HTML semantics without relying on model inference.
Tags and roles guide how element content and properties are compiled into an object's \textit{value}.
Nesting and references indicate how elements are related, providing the \textit{associations} needed to interpret that value.
For example, a dropdown menu offers users several options to choose from.
In HTML, the menu is defined by a \texttt{<select>} tag, with each choice specified by a nested \texttt{<option>} tag.
The compiler represents this menu as an action object whose \textit{value} specifies what the agent can do (select) and which options it can choose.
The context for this action comes from the text in a \texttt{<label>} tag, such as \textit{``clothing size''}.
The compiler includes this context in the object's \textit{association}, indicating what the options represent.
We implement these transformation rules as a finite set of \textit{constructors}, which the compiler applies while traversing the captured DOM tree in document order.
Constructors can recursively process nested elements, linking the resulting objects through references.
When an element has no dedicated constructor, the compiler processes its text content and recursively applies the same transformation rules to nested elements.
Text fragments separated by formatting tags are combined into a single text object.

Finally, \webcompiler organizes the resulting objects into a file tree following the page schema in \webcontract, preserving references between related objects.
The compiled page is registered in \webcache and assigned a \textit{page address}, allowing agents to access it in subsequent requests.

\vspace{-4pt}

\subsection{\webcache: Providing Persistent and Stateful Web Access}

\paragraph{Persistent Access.}
After receiving web search results, the interface fetches the top-ranked webpages, compiles them, and stores the compiled pages in \webcache.
Preparing these pages in advance provides three benefits.
First, the interface reports failed fetches (e.g., 404 errors or CAPTCHA challenges), so the agent does not need to spend additional turns identifying these access failures.
Second, the agent can get a quick overview of what each page offers without reading it in full.
The compiler provides a table of contents showing the available object groups and how to access them.
Third, operations support selective access to cached pages, allowing the agent to explore multiple webpages more efficiently and revisit them as the task progresses.

\vspace{-5pt}
\paragraph{Stateful Access.}
A core feature of \proposed is that \webcache tracks page state changes while preserving access to earlier states.
When the agent submits an \texttt{execute} request, the interface uses the connection retained during compilation to operate the corresponding control on the live webpage.
If the action changes the webpage, the interface compiles the updated page state and caches it under a new page address.
The action that produced the change is logged and reported to the agent in the response.
The earlier state remains read-only, so its objects can still be inspected even when the corresponding actions are no longer available on the live webpage.
This allows the agent to revisit past page states and review how the task progressed.
\vspace{-4pt}


%% file: Figures/main_figure.tex
\begin{figure*}[t]
    \centering
    \includegraphics[width=1\linewidth]{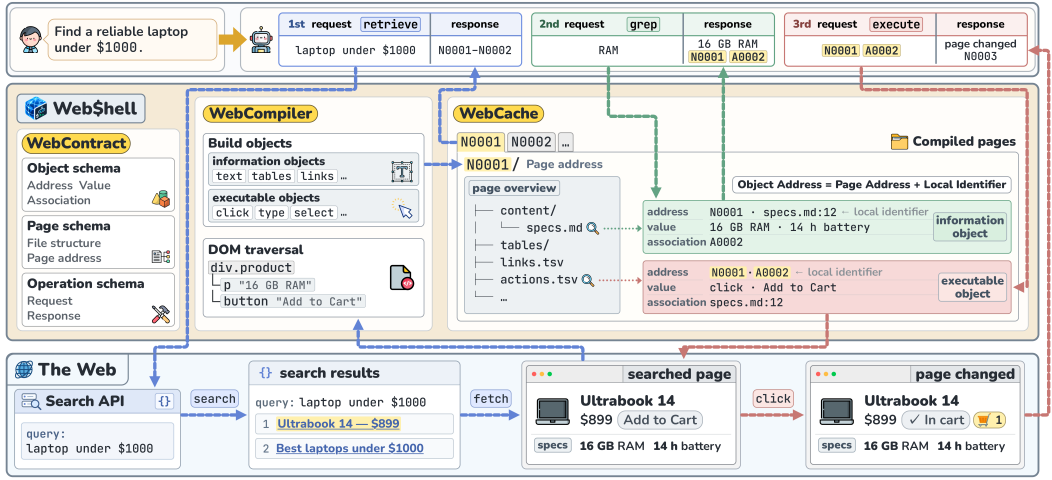}
    \vspace{-0.5cm} 
    \caption{Overview of \proposed. The agent retrieves webpages, selectively inspects cached objects, and executes associated actions through a unified request--response interface.}
    \label{fig:main_fig}
    \vspace{0.2cm}
\end{figure*}

%% file: MainText/050_experiments.tex
\paragraph{Benchmarks.}
\vspace{-4pt}
We evaluate \proposed across three categories of web tasks: web search, web navigation, and deep research.
Web search includes WideSearch~\citep{wong2026widesearch}, which requires collecting information across many sources into tables, and SealQA~\citep{pham2026sealqa}, which features challenging questions with conflicting search results.
For web navigation, we prioritize controlled evaluation settings.
WebArena-Verified~\citep{webarena-verified} provides simulated environments to evaluate task completion under reproducible conditions, while WebWalkerQA~\citep{wu2025webwalker} requires browsing within the same domain from a given starting URL.
For deep research, we cover research report writing in DeepResearchBench~\citep{du2026deepresearchbench} and quantitative analysis of evidence in DeepWeb-Bench~\citep{xie2026deepwebbench}.
For evaluation, we adopt the official metrics for each benchmark.
Further benchmark details are available in Appendix~\ref{appendix_benchmarks}.
\vspace{-4pt}

\paragraph{Baselines.}
We mainly compare \proposed with filesystem-based approaches, motivated by their competitive performance in our preliminary analysis (Section~\ref{sec:preliminary_analysis_2}).
We include:
(1) Fetch-then-Explore (FtE)~\citep{liu2026fetch}, which stores cleaned webpage text in files for selective reading and revisit;
(2) \hybridsbf, a hybrid configuration that directly combines search, browser, and filesystem access.
Additionally, we include (3) CLI-Anything Browser~\citep{yang2026clianything}, a software tool for shell-based browsing over accessibility trees, as a practical alternative to research baselines.

\vspace{-4pt}
\paragraph{Implementation Details.}
We use \texttt{gpt-5.6-luna} as the main-agent backbone with the reasoning effort set to \texttt{high}.
The turn budgets for Web Search, Web Navigation, and Deep Research are set to 100, 50, and 200, respectively.
For baselines, we follow their official implementations.
Experiments for each benchmark are run concurrently to reduce temporal variation in web search results.
We use the Serper API for web search\footnote{\url{https://serper.dev}}, and exclude HuggingFace pages containing reference answers to prevent answer leakage.
Further implementation details are provided in Appendix~\ref{appendix_implementation_details}.

\input{Tables/040_main_table}
\subsection{Main Results}
\label{sec:main_results}

\textbf{\proposed performs competitively across diverse web tasks.}
Table~\ref{tab:main_results} (Upper) compares \proposed with existing access models across Web Search, Web Navigation, and Deep Research.

On \textit{\textbf{Web Search}}, \proposed outperforms Fetch-then-Explore, the strongest baseline on both WideSearch and SealQA.
This advantage may partly stem from how \webcontract organizes content into common object types, such as body text, tables, and links, and defines consistent operations for accessing them across websites.
This matters because agents must locate and revisit evidence across sources while building an answer.
A common page organization gives them predictable locations to inspect, helping them direct subsequent requests to relevant content.

On \textit{\textbf{Web Navigation}}, \hybridsbf performs similarly to CLI-Anything Browser on WebArena-Verified, whereas \proposed achieves the best performance.
This comparison suggests that combining access capabilities alone does not necessarily improve task completion.
Navigation also requires agents to identify available actions and verify their outcomes.
\proposed supports these requirements by explicitly listing page controls and providing feedback on the effects of actions.
Specifically, the interface compares the previous and updated page states after each action and reports what has changed.
Together with the web search results, these gains suggest that a unified interface supports effective interaction while retaining the benefits of selective information access.

On \textit{\textbf{Deep Research}}, \proposed shows its clearest improvement on DeepWeb-Bench, which requires agents to gather numerical data from multiple sources and derive answers through multi-step reasoning.
When an evidence search (e.g., using \texttt{grep}) does not return the required information, agents may be uncertain about where to look next.
In such cases, \proposed's object association field may guide further inspection by providing references to related sections and tables, a role we examine further in Section~\ref{sec:analysis}.
Overall, \proposed's benefits are not limited to web search and navigation, but also extend to demanding deep research.

\textbf{\proposed achieves higher performance at lower cost.}
As shown in Table~\ref{tab:main_results} (Lower), the advantage is particularly evident in Deep Research, where agents must search and synthesize information from multiple sources.
Here, CLI-Anything Browser uses nearly three times as many turns as \proposed (52.4 vs.\ 17.6), while \hybridsbf uses nearly twice as many (30.9 vs.\ 17.6).
These results demonstrate the value of unified access for effective and cost-efficient agentic web access.
\vspace{-0.3cm}


%% file: Tables/040_main_table.tex
\definecolor{searchagent}{RGB}{242,237,250}
\definecolor{webagent}{RGB}{235,243,250}
\definecolor{filesystem}{RGB}{250,244,234}
\definecolor{webshell}{RGB}{237,247,240}
\setlength{\tabcolsep}{4pt}
\renewcommand{\arraystretch}{1.1}
\begin{table*}[t]
\caption{Performance (upper) and efficiency (lower) comparison on web search, web navigation, and deep research benchmarks. Efficiency reports average turns, tokens, and cost per task.}
\label{tab:main_results}
\centering
\vspace{3pt}
\scriptsize
\renewcommand{\arraystretch}{1.1}
\newlength{\agentW}\newlength{\tmpW}
\settowidth{\agentW}{\textbf{\proposed{} (ours)}}
\settowidth{\tmpW}{Fetch-then-Explore}
\ifdim\tmpW>\agentW \setlength{\agentW}{\tmpW}\fi

\setlength{\tabcolsep}{4pt}
\begin{tabularx}{\linewidth}{>{\raggedright\arraybackslash}p{\agentW}*{6}{>{\centering\arraybackslash}X}}
\toprule
\multirow{2}{*}{\textbf{Agent}}
& \multicolumn{2}{c}{\textbf{Web Search}}
& \multicolumn{2}{c}{\textbf{Web Navigation}}
& \multicolumn{2}{c}{\textbf{Deep Research}} \\
\cmidrule(lr){2-3} \cmidrule(lr){4-5} \cmidrule(lr){6-7}
& \makecell[t]{\textbf{WideSearch}\\\textit{Item F1} $\uparrow$}
& \makecell[t]{\textbf{SealQA}\\\textit{Acc.} $\uparrow$}
& \makecell[t]{\textbf{WA-Verified}\\\textit{SR} $\uparrow$}
& \makecell[t]{\textbf{WebWalker}\\\textit{Acc.} $\uparrow$}
& \makecell[t]{\textbf{DR-Bench}\\\textit{RACE Overall} $\uparrow$}
& \makecell[t]{\textbf{DW-Bench}\\\textit{Overall} $\uparrow$} \\
\midrule
\hybridsbf
& 0.738 & 0.646 & 0.578 & 0.590 & \underline{0.528} & 28.37 \\
Fetch-then-Explore
& \underline{0.745} & \underline{0.665} & 0.264 & \underline{0.636} & 0.527 & 28.27 \\
CLI-Any-Browser
& 0.660 & 0.650 & 0.575 & 0.588 & \underline{0.528} & 26.59 \\
\midrule
\rowcolor{webshell}
\textbf{\proposed{} (ours)}
& \textbf{0.782} & \textbf{0.720} & \textbf{0.679}
& \textbf{0.671} & \textbf{0.534} & \textbf{31.11} \\
\bottomrule
\end{tabularx}

\setlength{\tabcolsep}{3pt}
\begin{tabularx}{\linewidth}{>{\raggedright\arraybackslash}p{\agentW}*{9}{>{\centering\arraybackslash}X}}
\toprule
\multirow{2}{*}{\textbf{Agent}}
& \multicolumn{3}{c}{\textbf{Web Search}}
& \multicolumn{3}{c}{\textbf{Web Navigation}}
& \multicolumn{3}{c}{\textbf{Deep Research}} \\
\cmidrule(lr){2-4}\cmidrule(lr){5-7}\cmidrule(lr){8-10}
& \textbf{Turns} $\downarrow$ & \mbox{\textbf{Tok. (M)} $\downarrow$} & \textbf{Cost (\$)} $\downarrow$
& \textbf{Turns} $\downarrow$ & \mbox{\textbf{Tok. (M)} $\downarrow$} & \textbf{Cost (\$)} $\downarrow$
& \textbf{Turns} $\downarrow$ & \mbox{\textbf{Tok. (M)} $\downarrow$} & \textbf{Cost (\$)} $\downarrow$ \\
\midrule
\hybridsbf
& 28.3 & 1.84 & 0.070
& \underline{24.1} & \underline{0.58} & \underline{0.024}
& 30.9 & 2.75 & 0.111 \\
Fetch-then-Explore
& \underline{26.3} & \textbf{1.24} & \underline{0.063}
& 27.3 & 0.60 & 0.027
& \underline{27.8} & \underline{2.20} & \underline{0.097} \\
CLI-Any-Browser
& 47.4 & 3.08 & 0.102
& 25.6 & 1.15 & 0.044
& 52.4 & 4.03 & 0.137 \\
\midrule
\rowcolor{webshell}
\textbf{\proposed{} (ours)}
& \textbf{20.8} & \underline{1.51} & \textbf{0.062}
& \textbf{15.0} & \textbf{0.47} & \textbf{0.022}
& \textbf{17.6} & \textbf{1.87} & \textbf{0.096} \\
\bottomrule
\end{tabularx}
\vspace{2pt}
\end{table*}

%% file: MainText/060_analysis.tex
\vspace{-0.2cm}
Here, we analyze the effectiveness and robustness of \proposed.
See Appendix~\ref{app:analysis} for details.
\vspace{-0.2cm}

\input{Figures/mechanism_analysis/case_study}

%
%
\paragraph{How does \proposed help information seeking?}
\input{Figures/mechanism_analysis/2_analysis_evidence_figure}
\label{sec:analysis_evidence}
We observe that \proposed achieves the highest evidence coverage on both web search benchmarks, measured by how much gold evidence is present in the agent's context (Figure~\ref{fig:analysis_evidence}).
To understand how it gathers more evidence, we conduct a case study comparing \proposed with FtE, the strongest baseline on WideSearch (Figure~\ref{fig:case_study}, left).
The query asks for season award winners and their ratings across five football leagues.
A core distinction is that \proposed provides guidance for further inspection alongside the information it returns.
FtE's clean-text representation lists the \textit{``Premier League''} but omits the season results required to answer the query, leading to additional web searches.
In contrast, \proposed presents the league as an information object, with references to related season data in its association field.
These findings suggest that \proposed facilitates further exploration.
\vspace{-0.1cm}

%
%
\paragraph{How does \proposed help interaction?}
\label{sec:analysis_interaction}
A key feature of \proposed is that webpage controls are compiled into action objects with the context needed to interpret and use them.
Figure~\ref{fig:case_study} (right) illustrates this benefit in a task requiring an older article from a news archive.
The agent can either search for the article directly through the search form or use \textit{load more} to browse older entries.
\hybridsbf's observation, however, presents the form as an unlabeled text box and button, requiring the agent to reconstruct the context needed to understand and use it.
\hybridsbf instead repeatedly selects \textit{load more}, exhausting its turn budget without finding the article.
In contrast, \webcompiler preserves the label and query parameter in a self-contained action object, enabling \proposed to use the site search and reach the article.
This case illustrates how \proposed supports web navigation by reducing the need to infer how controls work.
\vspace{-0.3cm}
%

\paragraph{How does unified access to information and actions help?}
\input{Figures/mechanism_analysis/1_analysis_ablation}
\proposed presents information together with references to its associated actions, allowing agents to move directly from reading page content to interacting with it.
For example, a \texttt{grep} result for \textit{Set status} can provide the action reference needed to open the status dialog.
To examine the benefit of this connection, we remove these references and require agents to obtain available actions from a separate listing.
We observe that this separation consistently degrades performance and increases cost across the two web navigation benchmarks (Figure~\ref{fig:cross_reference_ablation}).
Connecting information and actions within the interface thus plays an important role in task completion.


%
%
\input{Figures/further_analysis/1_analysis_noise_figure}
\textbf{\proposed remains robust to irrelevant information.}
\label{sec:analysis_noise}
To test robustness to irrelevant information, we manually inject $50$, $100$, $200$, or $400$ distractor pages into \proposed's \webcache.
We find that \proposed maintains largely stable accuracy and inference costs even as the amount of irrelevant information increases substantially  (Figure~\ref{fig:analysis_noise}).
The baselines also maintain accuracy, but Fetch-then-Explore's cost more than doubles and \hybridsbf's rises steadily.
Fetch-then-Explore often lists the entire workspace before selecting what to read, so the agent processes more file names as pages accumulate.
\hybridsbf often avoids this step by searching a fetched page directly with \texttt{grep}, using the file path returned by \texttt{fetch}.
However, keeping track of these paths becomes harder as pages accumulate, and path errors sometimes lead the agent back to a full workspace listing.
In contrast, \proposed's page address system provides short references that agents can reuse in subsequent searches and reads, suggesting that reliable addressing helps selective access remain efficient as irrelevant content accumulates.

\vspace{-0.2cm}

\input{Figures/further_analysis/2_analysis_scaling_figure}
\paragraph{\proposed supports broader web access at a stable inference cost.}
\label{sec:analysis_scale}
We vary the number of prefetched pages per search ($k \in \{5,7,10\}$) on WideSearch to examine how \proposed handles a growing cache.
Performance varies modestly, while average inference cost remains nearly unchanged (Figure~\ref{fig:analysis_scaling}).
The agent continues to inspect a similarly sized subset even as more pages become available.
These results support an interface that keeps a broad collection of sources accessible while allowing the agent to determine the scope of inspection.
\proposed supports this flexibility without substantially increasing inference cost across the tested settings.


%% file: Figures/mechanism_analysis/case_study.tex
\begin{figure*}[t]
    \centering
    \includegraphics[width=0.99\linewidth]{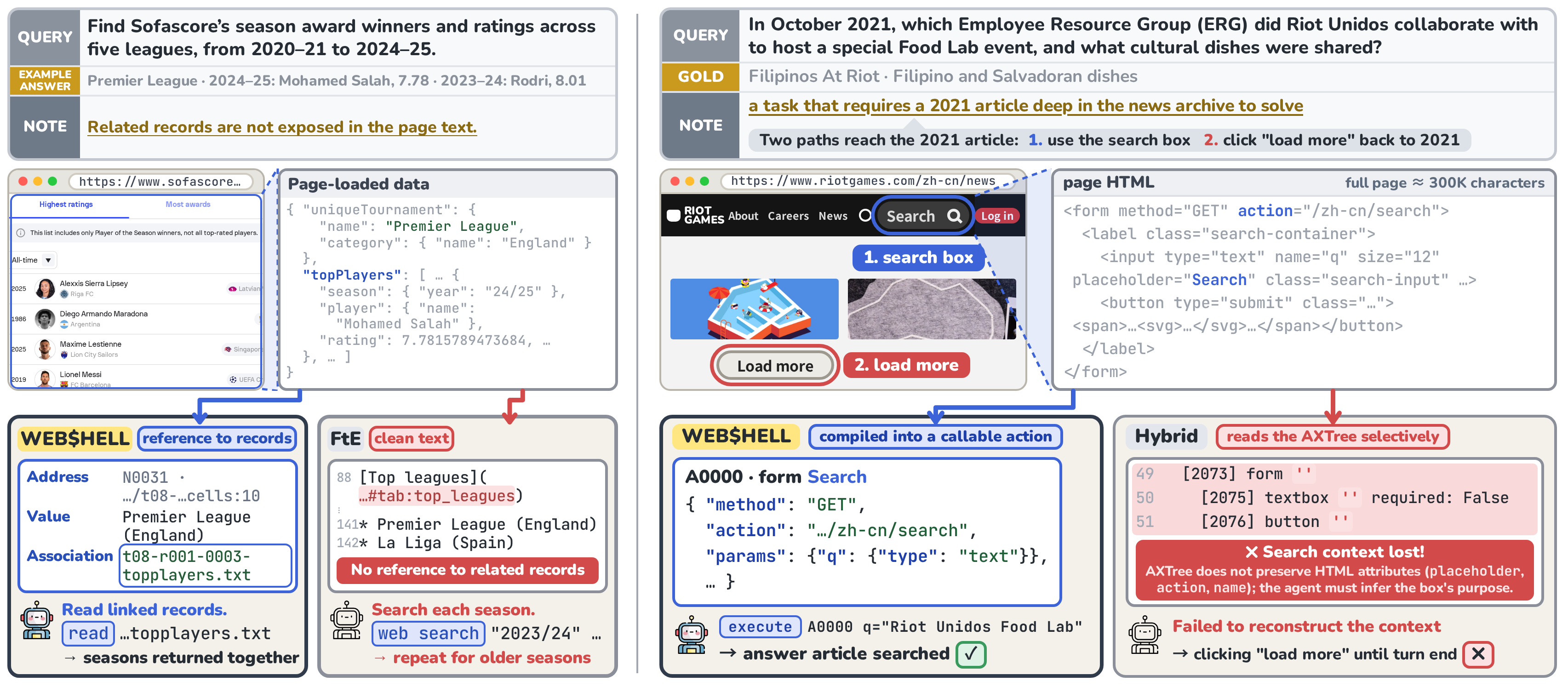}
    \caption{Case studies of \proposed vs.\ FtE in search (left) and \hybridsbf in navigation (right).}
    \label{fig:case_study}
\end{figure*}

%% file: Figures/mechanism_analysis/2_analysis_evidence_figure.tex
\begin{wrapfigure}{r}{0.49\textwidth}
    \centering    
    \vspace{-1.6em}
    \includegraphics[width=\linewidth]{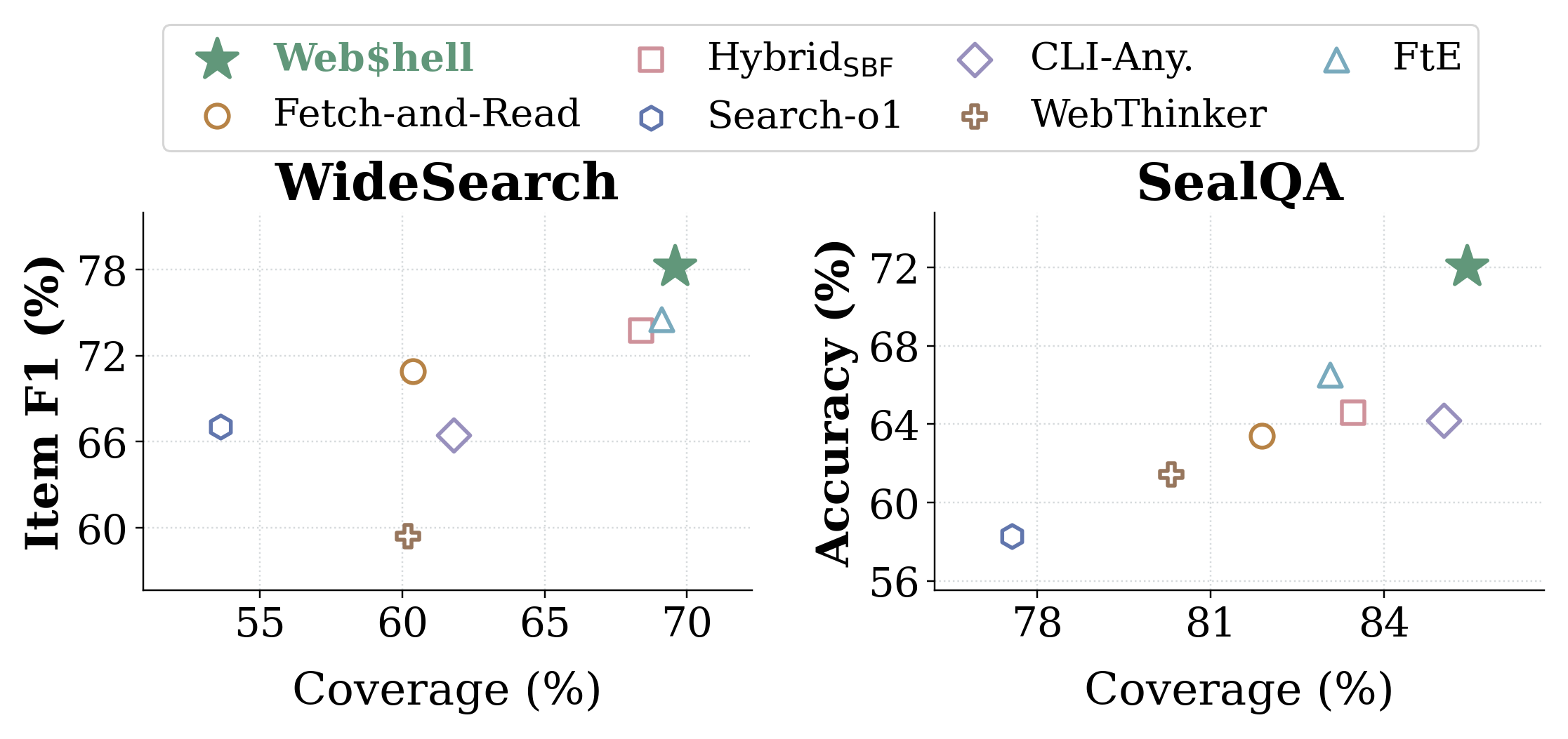}
    \vspace{-0.7cm}
    \caption{Evidence coverage and Accuracy on WideSearch and SealQA. Search-augmented agents are additionally included for comparison. }
    \label{fig:analysis_evidence}
    \vspace{-0.5cm}
\end{wrapfigure}

%% file: Figures/mechanism_analysis/1_analysis_ablation.tex
\begin{wrapfigure}{r}{0.4\textwidth}
    \centering
    \vspace{-1.4em}
    \includegraphics[width=\linewidth]
    {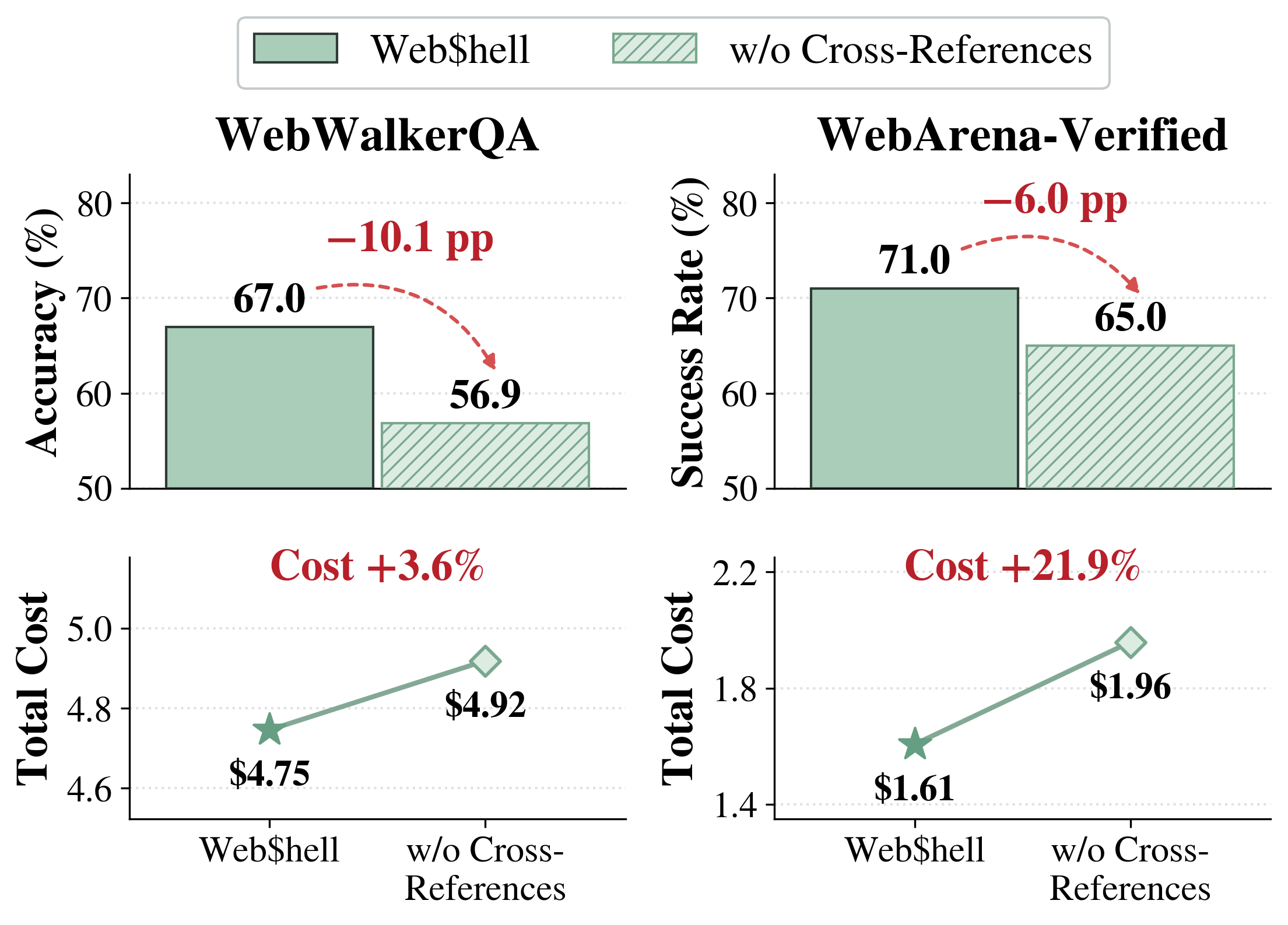}
    \vspace{-0.7cm}
    \caption{Effect of removing associations between information and actions.}
    \label{fig:cross_reference_ablation}
    \vspace{-0.5cm}
\end{wrapfigure}

%% file: Figures/further_analysis/1_analysis_noise_figure.tex
\begin{wrapfigure}{r}{0.51\textwidth}
    \centering
    \vspace{-1.4em}
    \includegraphics[width=\linewidth]{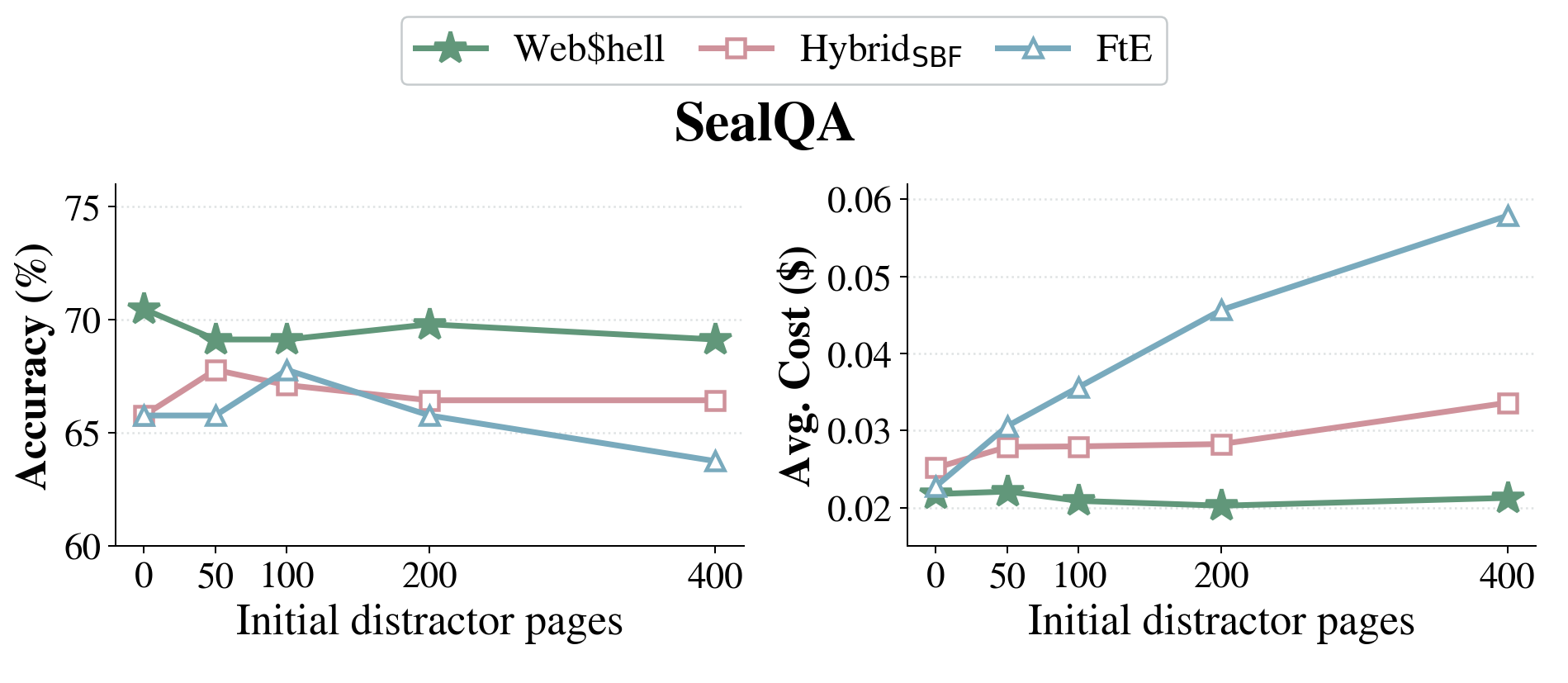}
    \vspace{-0.7cm}
    \caption{Effect of distractor pages on SealQA.}
    \label{fig:analysis_noise}
    \vspace{-0.3cm}
\end{wrapfigure}

%% file: Figures/further_analysis/2_analysis_scaling_figure.tex
\begin{wrapfigure}{r}{0.5\textwidth}
    \centering
    \vspace{-1.0em}
    \includegraphics[width=\linewidth]{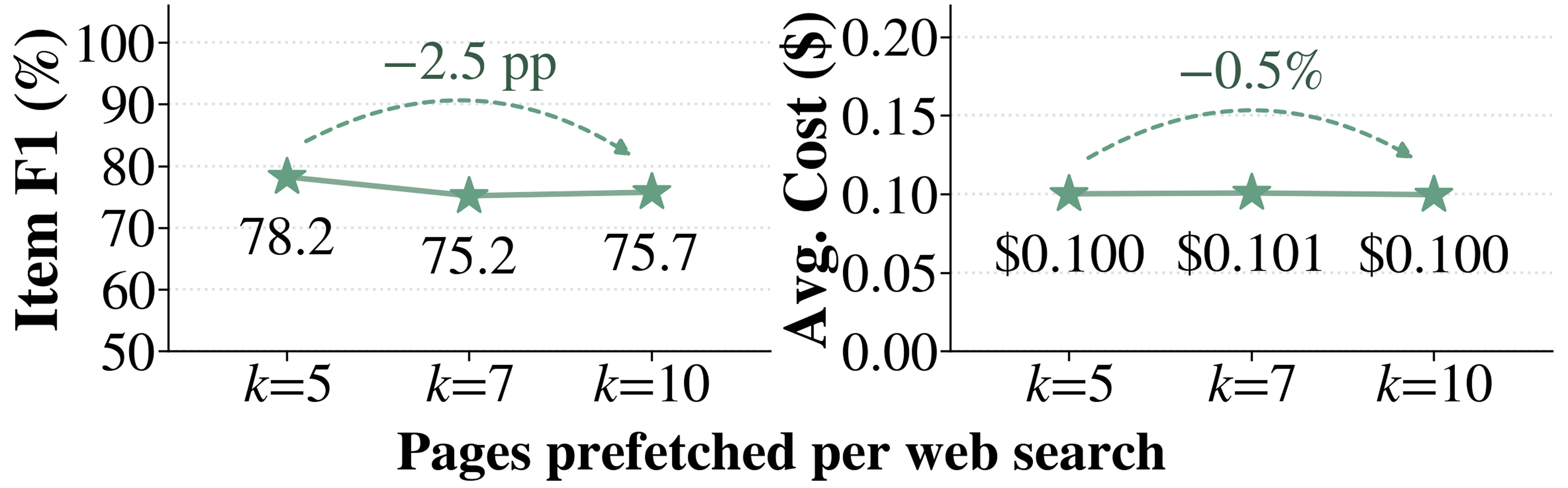}
    \vspace{-0.75cm}
    \caption{Effect of pages prefetched per search.}
    \label{fig:analysis_scaling}
    \vspace{-0.7em}
\end{wrapfigure}

%% file: MainText/070_conclusion.tex
This paper proposes \proposed, a unified interface for agentic web access.
It standardizes webpage information and actions as addressable objects, compiles live webpages into operable representations, and preserves them for selective and stateful access throughout a task.
Our experiments demonstrate strong performance and efficiency across web search, web navigation, and deep research, while further analyses show the benefits of connecting information with associated actions and maintaining persistent, addressable access as retrieved content accumulates.
We hope that \proposed will contribute to future work on more effective interfaces for the agentic web.

%% file: MainText/090_statements.tex
\subsection*{AI use statement}
LLMs were used to support manuscript writing and editing for clarity and readability, as well as to assist with code implementation for the experiments. All AI-assisted text was reviewed by the authors, and all AI-generated code was manually inspected and verified before being used in the experiments. The authors retain full responsibility for the final manuscript and experimental code, including any content produced with AI assistance.

%% file: MainText/080_appendix.tex
\clearpage

\section{Experimental Setup}
\label{appendix_experimental_details}

\subsection{Benchmarks and Metrics}
\label{appendix_benchmarks}

We evaluate on six benchmarks covering web search, web navigation, and deep research.
Together, they test information gathering across sources, interaction with websites, and synthesis of evidence into reports and quantitative answers.

\paragraph{Web Search.}
\begin{itemize}[leftmargin=1.5em,topsep=0pt,partopsep=0pt,itemsep=3pt,parsep=0pt]
\item \textbf{WideSearch}~\citep{wong2026widesearch} evaluates broad information gathering across webpages.
Agents find entities that meet a query's conditions and collect their attributes in a table.
The benchmark contains 100 English and 100 Chinese queries.
We report item-level F1 by comparing generated tables with reference tables.
We follow the benchmark's column-specific evaluation criteria, using rule-based checks for exact matches, numerical tolerances, dates, and URLs, and \texttt{gpt-5.6-luna} for columns requiring semantic judgments.
The same model also aligns differently worded entity names before row matching.
\item \textbf{SealQA}~\citep{pham2026sealqa} focuses on questions with conflicting, misleading, or incomplete search results.
Answering them requires resolving ambiguity or combining evidence across sources.
We evaluate all 254 questions in Seal-Hard and report answer accuracy.
\end{itemize}

\paragraph{Web Navigation.}
\begin{itemize}[leftmargin=1.5em,topsep=0pt,partopsep=0pt,itemsep=3pt,parsep=0pt]
\item \textbf{WebArena-Verified}~\citep{webarena-verified} provides reproducible, self-hosted websites for controlled evaluation of web interaction.
Tasks involve finding information, navigating between pages, and modifying content on shopping sites, discussion forums, and other web applications.
The benchmark contains 812 tasks, but 23 could not be reproduced because the required Wikipedia corpus could not be downloaded.
Our evaluation therefore covers 789 tasks.
We report task success rate using the benchmark's deterministic evaluator, which checks final responses and recorded interactions against task-specific completion criteria.
\item \textbf{WebWalkerQA}~\citep{wu2025webwalker} evaluates evidence gathering through navigation from a given starting URL.
Questions range from locating evidence on one subpage to combining information across multiple subpages.
We exclude questions whose reference pages are inaccessible or no longer contain the evidence for the gold answer after website updates.
We report answer accuracy on the remaining 420 questions.
\end{itemize}

\paragraph{Deep Research.}
\begin{itemize}[leftmargin=1.5em,topsep=0pt,partopsep=0pt,itemsep=3pt,parsep=0pt]
\item \textbf{DeepResearchBench}~\citep{du2026deepresearchbench} contains 100 tasks that require agents to synthesize web evidence into research reports.
We report RACE Overall, which compares generated and reference reports across comprehensiveness, insight/depth, instruction following, and readability.
An LLM judge applies task-specific criteria and weights to both reports.
The overall score ranges from 0 to 1, with 0.5 indicating equal scores for the generated and reference reports.
\item \textbf{DeepWeb-Bench}~\citep{xie2026deepwebbench} contains 100 tasks requiring quantitative analysis across sources.
Agents gather evidence and derive quantities to compare eight entities across eight analytical dimensions.
An LLM judge evaluates each cell against a reference answer and a task-specific rubric, which allows partial credit and specifies when information should be reported as unavailable.
We report the overall rubric score on a 0--100 scale.
\end{itemize}

\subsection{Baselines}
\label{appendix_baselines}

The preliminary comparison covers search-augmented, filesystem-based, and browser-based agents.
The main comparison includes Fetch-then-Explore, CLI-Anything Browser, and Hybrid\textsubscript{SBF}.

\paragraph{Search-augmented agents.}
\begin{itemize}[leftmargin=1.5em,topsep=0pt,partopsep=0pt,itemsep=3pt,parsep=0pt]
\item \textbf{Fetch-and-Read} follows a naive search-and-read approach, fetching retrieved webpages and passing their text directly to the agent. The agent reads the returned text and searches for further evidence as needed.
\item \textbf{Search-o1}~\citep{li2025searcho1} processes retrieved passages with its Reason-in-Documents module.
Given the search query and the reasoning so far, this module extracts relevant information and returns it to the main agent for further reasoning.
\item \textbf{WebThinker}~\citep{li2025webthinker} delegates information gathering to a Deep Web Explorer.
The explorer can issue further searches and open additional pages before returning its findings to the main agent.
\end{itemize}

\paragraph{Filesystem-based agents.}
\begin{itemize}[leftmargin=1.5em,topsep=0pt,partopsep=0pt,itemsep=3pt,parsep=0pt]
\item \textbf{Fetch-then-Explore}~\citep{liu2026fetch} saves fetched webpages as cleaned text in a persistent filesystem workspace.
The agent uses \texttt{grep} to locate relevant passages and \texttt{read} to inspect them, and can revisit saved pages later in the task.
\item \textbf{\mbox{CLI-Anything Browser}} is an open-source
tool for shell-based web interaction.
We include it to compare a browser-oriented tool with
\proposed's unified interface for information seeking
and interaction across webpages.
\end{itemize}

\paragraph{Browser-based agents.}
\begin{itemize}[leftmargin=1.5em,topsep=0pt,partopsep=0pt,itemsep=3pt,parsep=0pt]
\item \textbf{Agent-E}~\citep{agente} separates task planning from browser execution.
A planner assigns steps to a browser agent, which inspects page content and controls, performs actions, and reports the results to the planner.
\item \textbf{AgentOccam}~\citep{yang2025agentoccam} presents a simplified accessibility tree with identifiers for actionable elements.
The agent selects actions using the current observation and interaction history, and maintains a plan that it can branch or revise during navigation.
\item \textbf{BrowserUse}~\citep{browser_use2024} combines structured webpage observations with browser actions in an iterative agent loop.
It uses the current page and prior interaction history to select actions such as clicking, entering text, and navigating, then inspects the updated page.
\end{itemize}

\paragraph{Hybrid agents.}
\begin{itemize}[leftmargin=1.5em,topsep=0pt,partopsep=0pt,itemsep=3pt,parsep=0pt]
\item \textbf{Hybrid\textsubscript{SB}} combines web search and browser access.
It receives cleaned webpage text for information gathering and accessibility-tree observations for browser interaction.
\item \textbf{Hybrid\textsubscript{SBF}} adds a persistent filesystem workspace to Hybrid\textsubscript{SB}.
It stores these observations as files for selective inspection with shell operations.
\end{itemize}

\subsection{Evaluation Settings}
\label{appendix_evaluation_settings}
\label{appendix_implementation_details}

\paragraph{Implementation Details.}
We use \texttt{gpt-5.6-luna} as the main-agent model for all
methods, with reasoning effort set to \texttt{high}.
For established baselines, we use prompts from their original
implementations, adapted to the benchmark instructions and
available tools.
All methods receive the same main-agent turn budget within
each task category, with 100 turns for Web Search, 50 for
Web Navigation, and 200 for Deep Research.
We apply a 75-second page-access timeout and a 60-minute
execution limit per task across methods.
Methods with web-search access use
Serper\footnote{\url{https://serper.dev/}}.
For recoverable browser-loading failures, we fall
back to a direct HTTP request to retrieve the document.
Error pages and CAPTCHA challenges are reported as access
failures rather than treated as page evidence.
To prevent answer leakage, we exclude benchmark pages
containing reference answers, including those hosted on
Hugging Face, from search results and page access.

\paragraph{Answer evaluation.}
We use \texttt{gpt-5.6-luna} for LLM-based evaluation on WideSearch, SealQA, WebWalkerQA, and DeepResearchBench, and \texttt{gpt-5.5} on DeepWeb-Bench.
For SealQA and WebWalkerQA, the judge receives the question, final answer, and reference answer and checks whether the required information is correctly provided, allowing equivalent wording.
Each benchmark uses the same evaluation criteria and judge model across methods.

\section{Experimental Details}
\label{appendix_analysis_details}

\subsection{Preliminary Analysis}
\label{appendix_preliminary_analysis_details}
We randomly sample 100 tasks from each of WideSearch and
WebArena-Verified for the preliminary analysis.
Our comparison examines whether the designs developed for
information seeking remain effective when tasks also require
browser interaction.
For WebArena-Verified, we provide search-oriented agents with
the browser actions introduced in WebArena~\citep{Zhou2024webarena}, including
clicking, typing, scrolling, tab management, and page
navigation.
We preserve their clean-text observations because these
representations are integral to how the methods gather and
process web evidence.
Adding accessibility trees would address a different question:
whether extending those methods with browser-oriented
observations improves navigation.
We examine this extension through \hybridsb in Analysis II.

\paragraph{Per-call input and cost.}
We use per-call metrics to distinguish the context processed
at each model invocation from the number of invocations
needed to complete a task.
This distinction matters because some baselines distribute
work across a main agent and auxiliary planners or explorers.
We include all such calls and divide their total input tokens
and inference cost by the total number of LLM calls.
We report mean input tokens per call and inference cost per
100 calls, with the latter scaled for readability.
The resulting comparison captures the average inference
burden across each method's model calls, including work
delegated to auxiliary modules.

\subsection{Main Analysis}
\label{app:analysis}

\paragraph{Evidence coverage.}
We use evidence coverage to assess how effectively an interface
helps the agent access the information needed to answer a
question.
Using each benchmark's reference answers as the gold
information,
we check how much of that information appears in the tool
responses
exposed to the agent during task execution.
This distinction matters for interfaces that store webpages
for selective reading.
Relevant information may be present in the workspace without
ever reaching the agent's context.
We therefore count only the content actually returned to the
agent, including passages retrieved through \texttt{grep}
or \texttt{read}.
On WideSearch, coverage is the fraction of required reference-table
cells whose values appear in this content.
On SealQA, it is the fraction of questions whose reference
answer appears.
Each reference item is counted once, regardless of repeated
exposure.

\paragraph{Action-reference ablation.}
We examine which aspects of the unified interface contribute
to \proposed's effectiveness.
The preliminary analysis motivates reducing the need to
switch representations when moving from information gathering
to browser interaction.
This capability is supported jointly by \webcontract,
\webcompiler, and \webcache, which define, construct, and
retain objects with their associated actions.
Because these components operate on the same object schema,
removing an entire component would alter several aspects
of the interface together.
We therefore focus on the references linking information
objects to action objects, removing these references while
retaining both types of objects and \webcache.
Without these references, the agent must request a separate
control listing to obtain executable identifiers.
These identifiers remain valid until the page changes or
another listing is requested.
We compare task performance and LLM inference cost on
100 WebArena-Verified tasks and 100 WebWalkerQA questions.

\paragraph{Noise analysis.}
\label{appendix_noise_setup}
We examine whether agents can locate relevant information
efficiently when unrelated pages accumulate in their workspace.
For each SealQA run, we place $50$, $100$, $200$, or $400$
distractor pages in \proposed's initial \webcache and in the
filesystem workspaces of Fetch-then-Explore and \hybridsbf.
These pages are available for inspection but are not inserted
directly into the agent's context.
We measure answer accuracy and LLM inference cost at each
noise level.
We exclude CLI-Anything Browser because it exposes webpages
through open browser tabs.
Introducing distractors would therefore require additional
browser tabs (e.g., 400 chrome tabs), rather than stored pages in a workspace.
This experiment focuses on the latter setting.

\section{Implementation Details}
\label{appendix_method_details}

\subsection{\webcontract}
\label{appendix_webcontract}

\paragraph{Object schema.}
\label{appendix_object_schema}
\webcontract specifies how information and actions are addressed, interpreted, and accessed.
Every object has an \textit{address} identifying it within a captured page, a \textit{value} containing information or an action specification, and \textit{associations} providing context or references to related objects.
These properties are encoded in text, table, and action records according to the object's type.
For example, a table cell retains its row and column labels, while a selection action retains its purpose and available inputs.
Table~\ref{tab:appendix_object_schema} shows both objects on an illustrative clothing page.

\begin{table}[t]
\centering
\caption{Illustrative information and action objects from a
product options table.
Clicking the Medium cell selects that size.
The cell's association provides the corresponding action
address.
Addresses are simplified.}
\label{tab:appendix_object_schema}
\label{tab:appendix_compilation_example}
\small
\renewcommand{\arraystretch}{1.12}
\begin{tabularx}{\linewidth}{@{}l
>{\raggedright\arraybackslash}X
>{\raggedright\arraybackslash}X@{}}
\toprule
\textbf{Property}
& \textbf{Information object}
& \textbf{Action object} \\
\midrule
Address
& \texttt{tables/options.cells:12}
& \texttt{A0000} \\
\addlinespace[4pt]
Value
& \texttt{Medium}
& Click to select Medium. \\
\addlinespace[4pt]
Association
& Column: \textit{Size}.\newline
Related action: \texttt{A0000}.
& Table: \texttt{tables/options.cells}.\newline
Row: \textit{Medium}; column: \textit{Size}. \\
\bottomrule
\end{tabularx}
\end{table}

An object's full address combines its page address with a local identifier.
Text and table entries use a file path and line number; links, images, resources, and actions use identifiers such as \texttt{L0000}, \texttt{I0000}, \texttt{R0000}, and \texttt{A0000}.
These addresses remain valid for the stored capture even after the live page changes.

\paragraph{Page schema.}
\label{appendix_page_schema}
Each compiled page includes a table of contents that lists
its text sections and object groups, together with their
addresses for access.
Stored in \texttt{\_map.md}, this table of contents provides
an overview of the page layout shown in
Table~\ref{tab:appendix_page_layout}.
Text sections follow the webpage's heading structure and
include references to the tables and actions they contain.
The agent can therefore begin with the table of contents
or a search match, then follow these references to inspect
related objects.

\begin{table}[t]
\centering
\caption{Page schema. Representative files and the information preserved in their records.}
\label{tab:appendix_page_layout}
\small
\renewcommand{\arraystretch}{1.12}
\begin{tabularx}{\linewidth}{@{}>{\raggedright\arraybackslash}p{0.30\linewidth}>{\raggedright\arraybackslash}X@{}}
\toprule
\textbf{Files} & \textbf{Record content} \\
\midrule
\texttt{content/}, \texttt{boilerplate/} & Text sections and surrounding page content, with headings and references to associated objects. \\
\addlinespace[3pt]
\texttt{tables/} & Each cell stores its row identifier, row label, column label, and value. The table identifies its source. \\
\addlinespace[3pt]
\texttt{links.tsv} & Link identifier, destination, displayed text, and containing sections. \\
\addlinespace[3pt]
\texttt{images.tsv} & Image identifier, source URL, available description, and containing sections. \\
\addlinespace[3pt]
\texttt{actions.tsv} & Action identifier, type, label, context, input specification, and containing section. \\
\addlinespace[3pt]
\texttt{facts.kv} & Named values with keys identifying their source fields. \\
\addlinespace[3pt]
\texttt{resources.tsv} & Resource identifier, source URL, content type, and a reference to the stored response. \\
\addlinespace[3pt]
\texttt{long-values/} & Complete values stored separately, with their addresses and lengths retained in the parent records. \\
\bottomrule
\end{tabularx}
\end{table}

\paragraph{Operation schema.}
\label{appendix_operations}
A request specifies an operation, its arguments, and a page or object address when needed.
The response reports the outcome and returns the requested content or resulting page, together with addresses for subsequent requests.
Table~\ref{tab:appendix_operations} gives the request and response for each operation.
For the table cell in Table~\ref{tab:appendix_object_schema}, \texttt{lookup} with row \textit{Medium} and column \textit{Chest (cm)} returns \texttt{96} with its row context.
For the selection action, \texttt{execute} with page \texttt{N0000}, action \texttt{A0000}, and \texttt{size=m} selects \textit{Medium} and returns the resulting page address.

\begin{table}[t]
\centering
\caption{Requests and responses under \webcontract.}
\label{tab:appendix_operations}
\small
\renewcommand{\arraystretch}{1.12}
\begin{tabularx}{\linewidth}{@{}l >{\raggedright\arraybackslash}p{0.34\linewidth} >{\raggedright\arraybackslash}X@{}}
\toprule
\textbf{Operation} & \textbf{Request} & \textbf{Response} \\
\midrule
\texttt{retrieve} & A web search query. & Ranked results, cached page addresses, and page access outcomes. \\
\addlinespace[3pt]
\texttt{list} & No target required. & Cached page addresses, titles, and live-page status. \\
\addlinespace[3pt]
\texttt{grep} & A pattern, optionally restricted to a page or record group. & Matching content with its addresses. \\
\addlinespace[3pt]
\texttt{read} & A page or object address, optionally with a content range. & Requested content or page index, with a continuation position when needed. \\
\addlinespace[3pt]
\texttt{lookup} & A page or table address, with row and column criteria. & Matching cells with row context and associated actions. \\
\addlinespace[3pt]
\texttt{resolve} & A page address and an in-page reference. & The referenced content and its compiled address. \\
\addlinespace[3pt]
\texttt{execute} & A destination or action address, with any required input values. & The execution outcome and resulting page address. \\
\addlinespace[3pt]
\texttt{focus} & A page address. & The current state of the corresponding live page. \\
\bottomrule
\end{tabularx}
\end{table}

\subsection{\webcompiler}
\label{appendix_webcompiler}
\label{appendix_html_aria_semantics}

\paragraph{Compilation procedure.}
HTML describes a webpage's elements and their relationships, including headings, table cells, and input controls.
\webcompiler traverses the captured document in order and applies a constructor for each supported element type.
Constructors process nested elements recursively and preserve their relationships as associations.
Elements without a dedicated constructor contribute their text and nested content to the containing section.
The compiler then assigns file locations, connects references, and writes the page index without model inference.

\paragraph{Text and table construction.}
Headings divide text into sections, while formatting such as boldface remains part of the surrounding text.
Navigation, headers, and footers are stored separately when the page identifies these regions.
For tables, the compiler accounts for cells spanning multiple rows or columns and associates each value with its row and column labels.

\paragraph{Controls and forms.}
A control is an interactive element, such as a button, text field, or selection menu.
HTML and accessibility annotations called ARIA describe its type and label.
The compiler combines these declarations with input options, current values, and constraints such as required fields.
In the clothing example, the \texttt{<label>} references the \texttt{<select>} menu, connecting the purpose \textit{Clothing size} to the selection action.
Its nested \texttt{<option>} elements supply the displayed names and corresponding input values.
For a form, related fields and their submission control become one action.

\paragraph{Additional page data.}
A webpage may receive records that are not all displayed on screen.
The compiler also extracts structured data embedded in the page and text responses captured from the same origin during page loading.
It organizes these data into facts or tables, retaining references to their source and related records.

\paragraph{Long values.}
Individual fields can be much larger than the surrounding content; one vehicle-model field in a captured car-review page contains 1,767,349 characters.
Table and metadata values longer than 2,000 characters are stored separately, with their addresses and lengths retained in the parent records.
These large values can then be retrieved incrementally by requesting specific line or character ranges, while the complete value remains stored.

\subsection{\webcache}
\label{appendix_webcache}
\label{appendix_cache_states}

\paragraph{Page acquisition and reuse.}
\webcache stores pages obtained through web search or opened
directly by the agent, making them available throughout the
task.
For each search, the interface attempts to fetch the ranked
results and returns an address for each stored page, along with
any access failures.
The agent can then search or read the compiled content without
fetching the page again.
Browser tabs may also remain open, allowing the agent to return
to a page and interact with it after inspecting other pages.
These tabs share the browser session's cookies and local
storage, while each stored page state has its own address.

\paragraph{Live page states.}
After a control executes, the interface captures the resulting page.
Navigation to another URL produces a page capture; an update at the same URL produces a state capture when the compiled content changes.
For a state update, the response identifies the previous state and executed action, and lists changed, added, or removed files.
An unchanged result creates no new state, and a previously reached state reuses its cached address.
The action still executes even when its result is already cached.

\paragraph{Reading history and executing actions.}
As the agent navigates and interacts with a website, \webcache
preserves previously visited page states so that earlier
information remains available for reading.
The stored addresses identify this information, but accessing
them does not restore the browser to a previous state.
To distinguish stored records from pages that remain open,
\texttt{list} marks the most recently used live page as
\texttt{[active]} and other retained live pages as
\texttt{[live]}.
After an interaction changes a page, its tab is associated with
the resulting state, and \texttt{focus} selects a retained tab
and returns its current state.
Action execution depends on the browser state currently
available.
If a page must be reopened, the interface checks its current
content before attempting the requested action.
If the requested control cannot be confirmed, it returns the
current page information without executing the action.
If the tab for a saved interaction state is no longer
available, the response explains that the action cannot be
executed directly from that state.

\section{Additional Results and Analyses}
\label{appendix_additional_analyses}


\input{Figures/further_analysis/3_analysis_domains_figure}
\paragraph{Performance across domains.}
We examine whether the overall gains are concentrated in
particular domains.
As shown in Figure~\ref{fig:analysis_domains}, \proposed
matches or exceeds \hybridsbf across all categories in SealQA
and WebWalkerQA, and outperforms CLI-Anything Browser in each
category.
The comparison with Fetch-then-Explore reveals a more even
performance profile on WebWalkerQA.
Although Fetch-then-Explore achieves higher accuracy on
Education, its accuracy falls to 52.3\% on Game, whereas
\proposed remains between 63.5\% and 69.7\% across the four
categories.
Together, these results suggest that \proposed's effectiveness
extends across different subject areas and website categories,
rather than relying on exceptional performance in a narrow
domain.

\paragraph{Noise and selective access.}
Figure~\ref{fig:analysis_noise} shows that inference costs
respond differently as irrelevant pages accumulate, even though
all three methods support selective reading.
The trajectories suggest that this difference partly arises
from how agents manage file paths as the workspace grows.
Fetch-then-Explore typically starts with a full workspace
listing, whose length increases with the number of stored
pages.
\hybridsbf often avoids this step by passing the file path
returned by \texttt{fetch} directly to \texttt{grep}.
As more pages accumulate, however, keeping track of the
relevant paths becomes more difficult, and we observe cases
where incorrectly reproduced paths lead to full workspace
listings.
\proposed provides short addresses for both pages and
individual objects, helping agents specify subsequent searches
and reads more reliably.
These observations suggest that efficient selective access
requires manageable references to stored content, so that
locating evidence does not become increasingly costly as the
workspace grows.

\paragraph{Scaling and evidence retention.}
\label{appendix_scaling_analysis}
Figure~\ref{fig:analysis_scaling} examines how \proposed
handles a growing collection of prefetched pages.
By storing pages in \webcache and exposing them through
selective operations, \proposed makes more information
available without requiring the agent to read every page.
The agent inspects a similarly sized subset as the collection
grows, with the fraction of cached pages inspected decreasing
from 14.7\% at $k=5$ to 10.4\% at $k=10$.
Yet unaccessed content can still contain useful evidence.
At $k=10$, 9.5\% of the gold information in \webcache has not
reached the agent's observations.
Retaining this content preserves the opportunity to retrieve it
through later requests.
This highlights the value of separating information
availability from immediate consumption, while motivating cache
management that preserves potentially useful evidence.

\begin{table}[htbp]
\centering
\begin{minipage}{0.46\linewidth}
\caption{Mean compilation time per page for \proposed.}
\label{tab:appendix_interface_overhead}
\small
\begin{tabularx}{\linewidth}{@{}Xr@{}}
\toprule
\textbf{Benchmark} & \textbf{Time (s)} \\
\midrule
WideSearch & 1.62 \\
SealQA & 2.02 \\
WebArena-Verified & 0.98 \\
WebWalkerQA & 0.46 \\
DeepResearchBench & 1.38 \\
DeepWeb-Bench & 2.79 \\
\bottomrule
\end{tabularx}
\end{minipage}
\end{table}

\paragraph{Compilation overhead.}
\label{appendix_interface_overhead}
\proposed and the main baselines share the same web-search API.
However, page loading time depends heavily on the websites
accessed and their rendering behavior, making total elapsed
time an indirect measure of interface overhead.
\proposed also supports fallback access when retrieval fails
and closes failed pages to avoid unnecessary continued
processing.
We therefore measure compilation overhead directly as the time
required to compile acquired content and store the resulting
objects.
Across benchmarks, this takes 0.46--2.79 seconds per page on
average (Table~\ref{tab:appendix_interface_overhead}),
indicating a modest per-page compilation cost.

\paragraph{Cost comparison with Fetch-then-Explore.}
\proposed achieves higher performance than Fetch-then-Explore
across Web Search and Deep Research with fewer model calls
and comparable inference cost per task
(Table~\ref{tab:main_results}).
To understand this balance, we examine the cumulative token
usage per task (Table~\ref{tab:appendix_token_breakdown}).
On Web Search, \proposed uses fewer uncached input and output
tokens, although its cached input is higher.
The input totals include instructions explaining how to use
each interface and its tool definitions.
Under the evaluated pricing, cached input costs 10\% of
regular input, keeping the total inference cost comparable
despite \proposed's higher total token count.
Thus, \proposed achieves stronger information-seeking
performance within a similar inference budget, while
additionally supporting browser interaction.

\begin{table}[h]
\caption{Mean token usage (k) and LLM inference cost (USD) per
task.
Cached input refers to the model provider's prompt cache.
Each task category averages its two benchmark means.
Lower values in each column are bold.}
\label{tab:appendix_token_breakdown}
\centering
\vspace{3pt}
\scriptsize
\setlength{\tabcolsep}{4pt}
\renewcommand{\arraystretch}{1.1}
\begin{tabularx}{\linewidth}{
  @{}l*{8}{>{\centering\arraybackslash}X}@{}
}
\toprule
\multirow{2}{*}{\textbf{Agent}}
& \multicolumn{4}{c}{\textbf{Web Search}}
& \multicolumn{4}{c}{\textbf{Deep Research}} \\
\cmidrule(lr){2-5}
\cmidrule(lr){6-9}
& \textbf{Cached}
& \textbf{Uncached}
& \textbf{Output}
& \textbf{Cost}
& \textbf{Cached}
& \textbf{Uncached}
& \textbf{Output}
& \textbf{Cost} \\
\midrule
Fetch-then-Explore
& \textbf{1,120.5} & 110.5 & 10.0 & 0.0626
& 2,059.8 & \textbf{120.6} & 21.2 & 0.0969 \\
\rowcolor{webshell}
\textbf{\proposed}
& 1,410.3 & \textbf{87.2} & \textbf{9.1} & \textbf{0.0623}
& \textbf{1,705.9} & 141.3 & \textbf{20.3} & \textbf{0.0957} \\
\bottomrule
\end{tabularx}
\end{table}

%% file: Figures/further_analysis/3_analysis_domains_figure.tex
\begin{wrapfigure}{r}{0.4\textwidth}
    \centering
    \includegraphics[width=\linewidth]{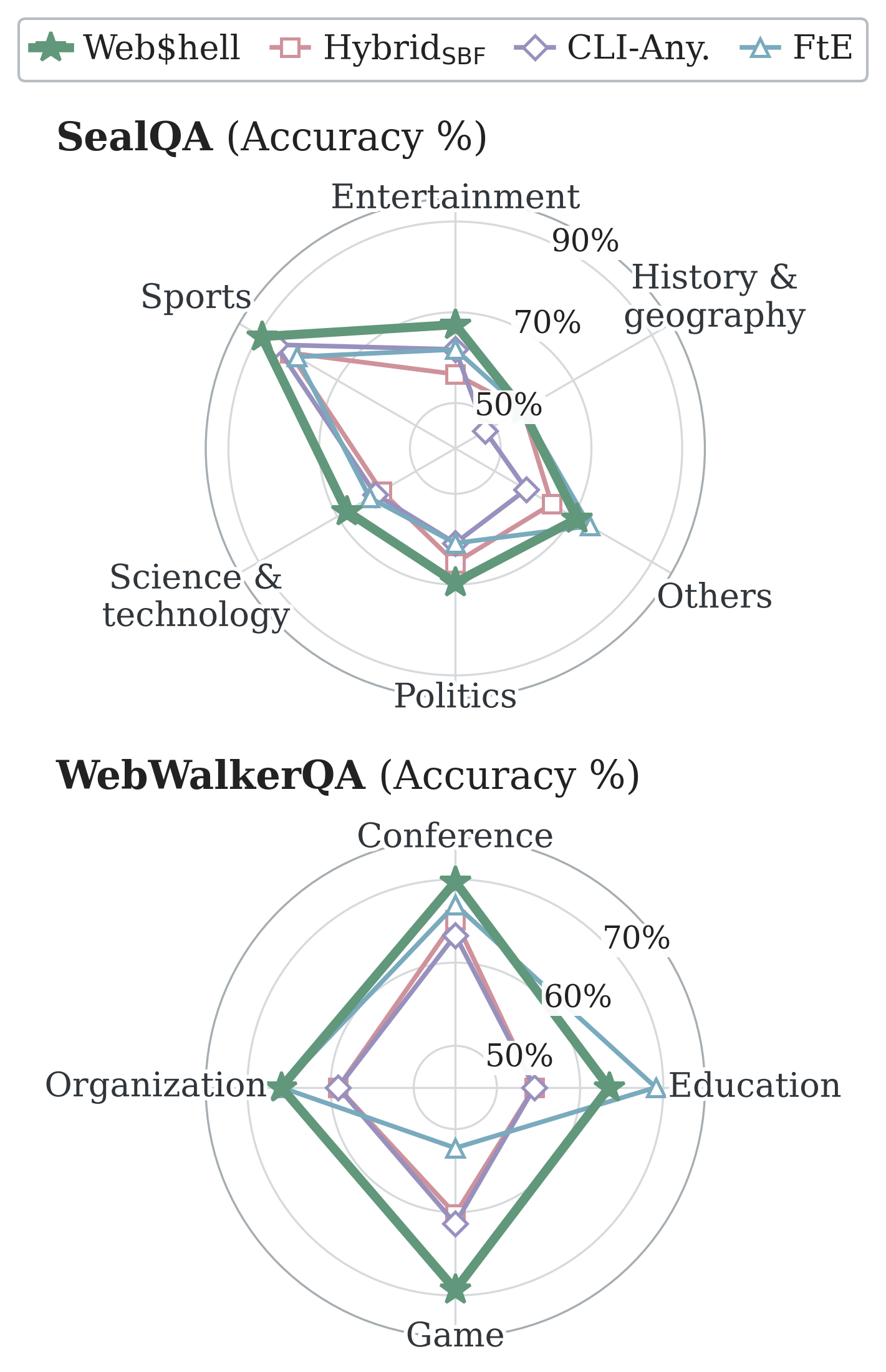}
    \caption{Domain analysis}
    \label{fig:analysis_domains}
\end{wrapfigure}